\pdfoutput=1

\documentclass[11pt]{article}

\usepackage[]{acl}

\usepackage{times}
\usepackage{latexsym}
\usepackage{amsmath}

\usepackage{multirow}
\usepackage{booktabs}

\usepackage{graphicx}
\usepackage{enumerate,enumitem}

\usepackage[T1]{fontenc}

\usepackage[utf8]{inputenc}

\usepackage{microtype}

\usepackage{inconsolata}

\title{Revealing the Parametric Knowledge of Language Models:\\ A Unified Framework for Attribution Methods}

\newcommand{\nanew}{NA-Instances}
\newcommand{\ianew}{IA-Neurons}

\author{Haeun Yu \quad Pepa Atanasova \quad Isabelle Augenstein \\
  University of Copenhagen \\
  {{\tt \href{mailto:hayu@di.ku.dk}{hayu@di.ku.dk}}
  \quad {\tt \href{mailto:pepa@di.ku.dk}{pepa@di.ku.dk}}
  \quad {\tt \href{mailto:augenstein@di.ku.dk}{augenstein@di.ku.dk}}}
  }

\begin{document}
\maketitle
\begin{abstract}

Language Models (LMs) acquire parametric knowledge from their training process, embedding it within their weights. The increasing scalability of LMs, however, poses significant challenges for understanding a model's inner workings and further for updating or correcting this embedded knowledge without the significant cost of retraining. This underscores the importance of unveiling exactly what knowledge is stored and its association with specific model components. Instance Attribution (IA) and Neuron Attribution (NA) offer insights into this training-acquired knowledge, though they have not been compared systematically. Our study introduces a novel evaluation framework to quantify and compare the knowledge revealed by IA and NA. To align the results of the methods we introduce the attribution method \nanew\ to apply NA for retrieving influential training instances, and \ianew\ to discover important neurons of influential instances discovered by IA. We further propose a comprehensive list of faithfulness tests to evaluate the comprehensiveness and sufficiency of the explanations provided by both methods. 
Through extensive experiments and analysis, we demonstrate that NA generally reveals more diverse and comprehensive information regarding the LM's parametric knowledge compared to IA. Nevertheless, IA provides unique and valuable insights into the LM’s parametric knowledge, which are not revealed by NA. Our findings further suggest the potential of a synergistic approach of combining the diverse findings of IA and NA for a more holistic understanding of an LM's parametric knowledge.

\end{abstract}

\section{Introduction}

\begin{figure}[t]
\includegraphics[width=1.0\columnwidth]{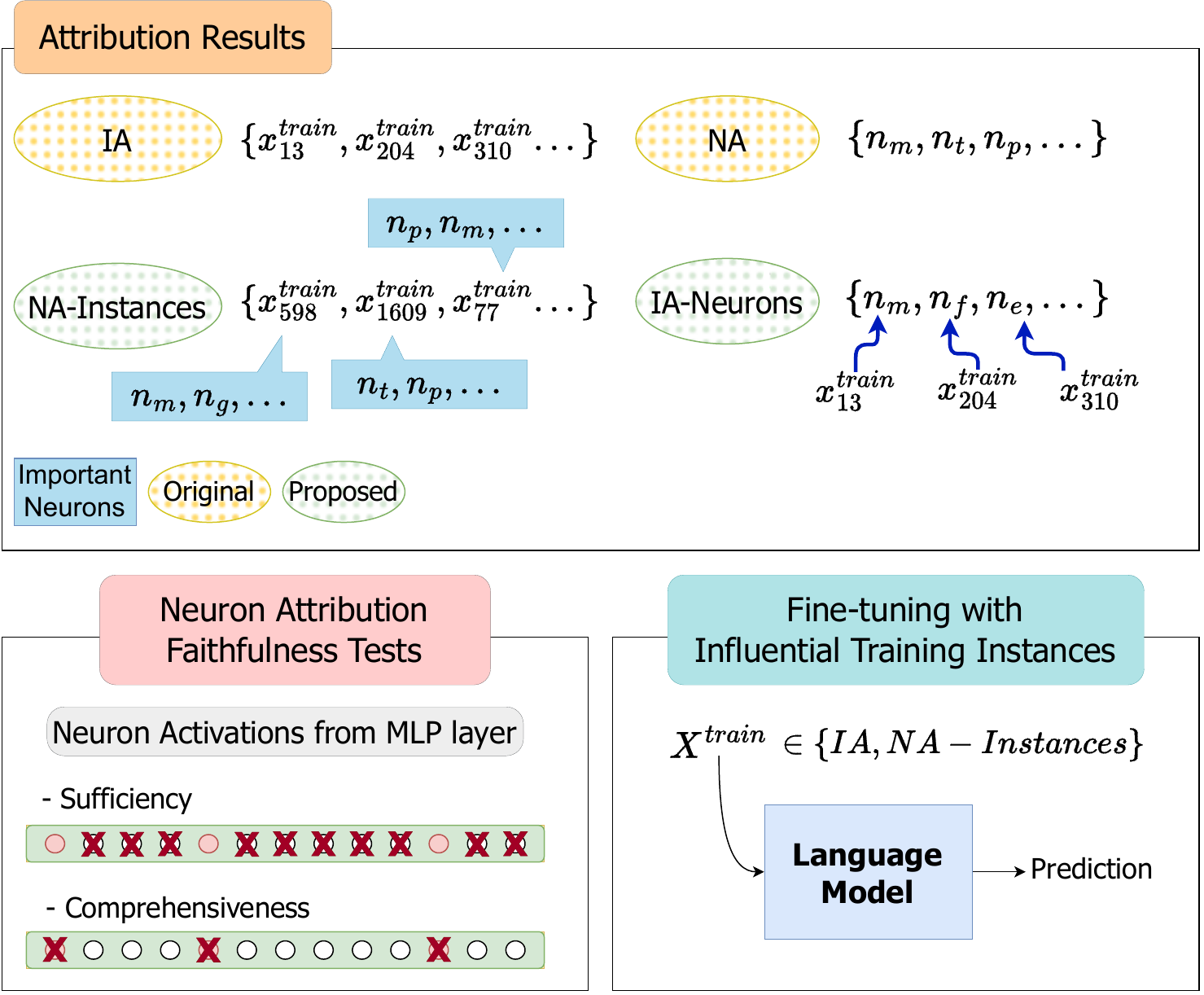}
\caption{Proposed evaluation framework comparing Instance and Neuron Attribution methods by examining most influential training instances ${x_i^{train}}$ and most important neurons ${n_m, n_t, n_p, \dots}$. Tests for Sufficiency (activation of key neurons) and Completeness (suppression of the activation of key neurons) -- bottom left,  alongside fine-tuning with influential training instances -- bottom right, assess the attribution methods' fidelity to the model's mechanisms.}
\label{fig1}
\end{figure}

Language Models encode the knowledge acquired during training as numeric values within the models' weights, transforming raw information from the training dataset into structured, internal representations. This embedding of knowledge, while fundamental to an LM's functionality, renders the model’s inner workings opaque. To unravel the internal mechanisms of an LM and investigate the parametric knowledge encoded in an LM's weights, the development of eXplainable AI (XAI) methods is paramount.  

Research in this domain has explored LM’s parametric knowledge with various attribution methods \cite{bejan-etal-2023-make, fan2023evaluating}. Commonly used among these are \textit{Instance Attribution} (IA, \citet{pmlr-v70-koh17a,inputsim-nips-2019,tracin-nips-2020}) and \textit{Neuron Attribution} (NA, \citet{dai-etal-2022-knowledge,meng2022locating}). IA identifies training instances that influence the model’s parametric knowledge leveraged for its prediction on a test instance. However, despite their utility, IA methods have been criticized for their sensitivity to the choice of hyperparameters and for producing highly homogenous results \cite{pezeshkpour-etal-2021-empirical}. On the other hand, NA locates specific neurons that hold the most important parametric knowledge for prediction on a test instance. While NA has proven valuable in locating and editing the structured knowledge within a model \cite{dai-etal-2022-knowledge, meng2022locating}, the granular nature of the neuron analysis presents challenges in the interpretation of the outcomes, necessitating manually defined human concepts for interpretation \cite{sajjad-etal-2022-neuron}. 

While IA and NA present different views on the parametric knowledge employed in an LM's prediction, no existing work has contrasted the two techniques to determine if they offer similar or complementary insights that could lead to a more comprehensive understanding of an LM's inner workings. This paper seeks to bridge this gap by establishing a \textit{unified evaluation framework}, illustrated in Figure \ref{fig1}, that allows for the comparison of these disparate attribution methods, particularly focusing on their application for autoregressive LMs. 

\textbf{Aligning the Results of Attribution Methods.} To allow for comparing the results obtained with NA to those obtained with IA, we first introduce \textit{a novel attribution method \nanew}\ (bottom left, `Attribution Results', Figure \ref{fig1}; \S\ref{sec:test:ia}) that retrieves the training instances sharing the most similar neuron activations for each test instance. \nanew\ allows one to interpret the granular NA results and to compare its results to IA. On the other hand, to align the results of IA with NA, we introduce \textit{\ianew}\ (bottom right,  `Attribution Results', Figure \ref{fig1}; \S\ref{sec:test:na}), which finds the neurons of the most important training instances discovered by IA that have the highest activation.

\textbf{Unified Evaluation Framework for Attribution Methods.} With the aligned results from IA and NA, we introduce two evaluation schemes. Interpretation of the attribution results is important, but it is also crucial to check whether the given explanation fully represents the model's underlying behavior \cite{Ross2017Right4right}. Therefore, firstly, \textit{faithfulness tests} are designed to assess whether the \textit{neurons discovered by each method are both sufficient and comprehensive} in representing the parametric knowledge used by the model for a prediction (bottom left, Figure \ref{fig1};\S\ref{sec:test:na}). Note that previous work has only performed the comprehensiveness test for NA \cite{dai-etal-2022-knowledge}, while such evaluations have not been extended to IA. We find that most neurons can be removed without a significant number of changed predictions compared to the original model, revealing that much of the model's knowledge is not stored in neurons located in the Multilayer Perceptron Layers (MLP). 
Secondly, we conduct \textit{fine-tuning with a varying number of influential training instances} discovered by the two IA methods (bottom right, Figure \ref{fig1};\S\ref{sec:test:ia}) to assess how sufficient the training instances are in representing the parametric knowledge used by the model for a prediction. We find that IA and \nanew\ perform on par in both finding sufficient and comprehensive neurons and sufficient influential training instances. In addition to this, results show that influential training instances obtained by IA methods do not yield a better-fine-tuned model than randomly selected training instances.

\textbf{Characterising Differences between Attribution Methods.} We conclude with an extensive analysis (\S\ref{sec:analysis}) of the characteristics of the attribution methods focusing on their diversity and utility. By comparing the group of selected instances and neurons (\S\ref{sec:analysis:1}), we first observe that the instances discovered by NA-Instances and IA have a very small overlap.
In contrast, the group of neurons from IA-Neurons highly overlaps with the group of neurons from NA, where the latter discovers a large group of neurons in addition to the overlapping ones.
Furthermore, from the analysis (\S\ref{sec:analysis:2}) and results from \S\ref{sec:finetuningtest}, we find that the more diverse the group of influential training instances, the better the performance of the model fine-tuned with the group. Finally, results from \S\ref{sec:analysis:3} show that NA-Instances performs better at discovering dataset artifacts than IA methods. Overall, our findings call for future work incorporating both IA and NA methods for a better understanding of LM's parametric knowledge.

\section{Related Work}
To understand how models encode and utilise parametric knowledge, researchers have employed a variety of approaches \cite{xiong2024explainable}. Among these, Instance Attribution and Neuron Attribution have emerged as two principal methodologies.

\textbf{Instance Attribution (IA).}
IA identifies the training instances most influential to a model's prediction for a given test instance. IA's primary strength is that it provides a human interpretable explanation of the model's encoded parametric knowledge. \citet{pmlr-v70-koh17a} first introduced the Influence Function (IF), a method to calculate the impact of each training instance on a model's prediction, utilizing gradients and Hessian-vector products. Additional methods in this domain include Input Similarity \cite{inputsim-nips-2019}, and TracIn \cite{tracin-nips-2020}, which explore similar concepts from different perspectives. \citet{inputsim-nips-2019} defined the notion of similarity of inputs from the neuron network perspective and applied the method to retrieve the training instances close to the test instance. Further, TracIn proposed to track an individual loss of a test instance by looking at the moment when a certain training instance is introduced. We note that most IA methods are developed upon image datasets and models, rather than NLP-specific downstream tasks. This calls for the development of IA methods specialized in NLP tasks, especially in the era of LMs. From an NLP perspective, \citet{pezeshkpour-etal-2022-combining} combined IA methods with feature attribution methods to identify the part of the training instance that influences test prediction. Notably, \citet{han-etal-2020-explaining} applied IF across various NLP tasks, revealing its potential to detect and correct training data-induced biases. 

\textbf{Neuron Attribution (NA).}
\citet{dai-etal-2022-knowledge} first coined the term ``knowledge neuron'' referring to the medium where the knowledge lies within the model. Their work illustrated that the factual knowledge encoded within a model's parameters could be altered by manipulating these specific neurons. Following this, \citet{meng2022locating} refined the process of locating knowledge within LMs through causal tracing, demonstrating that feed-forward layers in the model's mid-blocks play a crucial role in encoding factual knowledge. On the contrary, \citet{hase2023does} claimed that the knowledge neurons discovered by the causal tracing method often do not match the actual model weights that can modify the particular knowledge. This ongoing debate highlights the lack of consensus regarding the utility of NA.

\textbf{Evaluation of Attribution Methods.} Several works have assessed the utility across attribution methods of the same group. Notably, a recent investigation into Instance Attribution \cite{gu-etal-2023-iaeval} examines three distinct methods, evaluating them with four metrics -- sufficiency, completeness, stability and plausibility. Regarding NA, \citet{fan2023evaluating} introduced an evaluation framework and compared six different NA methods. They assessed the NA methods with two compatibility metrics which are inspired by voting theory. One is called AvgOverlap and the other is called NeuronVote. They return one method's compatibility score with respect to other methods using a ranked list of important neurons. Despite these advancements, there remains a gap in the literature regarding a direct comparison between IA and NA methods, which are both used to reveal the parametric knowledge used by an LM. This paper aims to bridge this research gap by presenting a unified framework that allows for a comprehensive comparative analysis of the two methods.

\section{An Evaluation Framework for Attribution Methods} \label{sec:3}

We next describe the proposed unified evaluation framework for attribution methods. 
To allow for the comparison of the two attribution methods IA and NA, we first propose to align their results to common views over the LM's parametric knowledge (\S\ref{sec:method:align}). Then, 
we describe the tests employed in the evaluation framework to assess whether the uncovered methods reflect the reasons employed by the models in their predictions (\S\ref{sec:test:na}, \S\ref{sec:test:ia}).

\subsection{Preliminaries}
Consider a test dataset $X^{Test}=[x_1, x_2, \dots, x_n]$, consisting of $n$ test instances and a train dataset $X^{Train}=[x_1^{train}, x_2^{train}, \dots , x_m^{train}]$ consisting of $m$ training instances. Consider also an LM $f$ used to make a prediction on an instance: $f(x) = \widehat{y}$. Furthermore, we introduce an attribution method $\mathcal{A} \in \{IA, NA, \ianew, \nanew\}$. For each test instance $x_i$, and a model making a prediction $f(x_i)$, $\mathcal{A}$ computes influence scores of the training instances $\mathcal{S}_{f(x_i)}=[s_{i, 1}^{train}, s_{i, 2}^{train}, \dots, s_{i,m}^{train}]$ designating the extent of influence of the corresponding training instance on the learned parametric knowledge of an LM employed in its prediction on $x_i$. $\mathcal{A}$ also returns $r$ most important neurons  $\mathcal{N}_{f(x_i)}=[n_i^1, n_i^2, \dots, n_i^r]$ and their corresponding attribution scores $\mathcal{NS}_{f(x_i)}=[ns_i^1, ns_i^2, \dots, ns_i^r]$ which are regarded as important for the model's prediction on $x_i$. 

\subsection{Aligning the Results of Attribution Methods.} 
\label{sec:method:align}
First, to compare the results of IA to the list of most important neurons discovered by NA, we introduce \textbf{\ianew}. \ianew\ first produces a list of scores for the influence of the training instances $\mathcal{S}_{f(x_i)}$ and takes $r$ most influential of them: $[x^{train}_{a}, x^{train}_{b}, \dots, x^{train}_{r}]$. Then, for each instance in the list, the NA method collects the most important neuron, arriving at the list of top-1 important neurons of the $r$ most influential training instances for a prediction: $\mathcal{S \circ N}_{f(x_i)}=[n_{x^{train}_a}^1, n_{x^{train}_b}^1, ..., n_{x^{train}_r}^1]$. 

Second, to compare the NA results to the influence scores of the training instances, we introduce \textbf{\nanew}. To design \nanew, we propose \textit{Discounted Cumulative Neuron Similarity} (DCNS) based on the Discounted Cumulative Gain evaluation measure \cite{dcg2002acm} used for the evaluation of ranked results in Information Retrieval. It takes both the rank of the retrieved item and its relevance score into account. Inspired by this, we utilise DCG to score training instances based on their neuron activation similarity with a given test instance. 
Thus, for a neuron that is important for the test instance prediction, we consider both the importance rank and the attribution score of the neuron from a training instance prediction. Equation \ref{eq1} shows the scoring function that takes the test instance $x_i$'s important neuron list $\mathcal{N}_{f(x_i)}=[n_{i}^1, n_{i}^2, \dots, n_{i}^r]$, the train instance $x_t$'s important neuron list $\mathcal{N}_{f(x_t)}=[n_{t}^1, n_{t}^2, \dots, n_{t}^r]$ and its attribution scores $\mathcal{NS}_{f(x_t)}=[ns_t^1, ns_t^2, ..., ns_t^r]$. It calculates the influence score $g_{i, t}^{train}$ of training instance $x_t$ for the prediction of test instance $x_i$.
\begin{equation}
    g_{i, t}^{train} = \sum_{m=[1,r]; n_t^m \in \mathcal{N}_{f(x_i)}} \frac{2^{ns_t^m}-1}{log_2(m+1)}
\label{eq1}
\end{equation}

Similar to the DCG metric, we penalize the attribution score of the common important neuron with its rank on the list of important neurons for the training instance.
In summary, a particular training instance is influential for a test instance, when the important neurons of the test instance are highly ranked for the training instance.

\subsection{Neuron Attribution Faithfulness Tests}
\label{sec:test:na}

We employ two faithfulness tests to assess which Neuron Attribution method returns a list of most important neurons $\mathcal{N}_{f(x_i)}$ for an LM's prediction on a test instance $x_i$ that is more faithful to the parametric knowledge employed by the LM in its prediction. Sufficiency and Comprehensiveness are two representative evaluation metrics of faithfulness. To evaluate \textbf{Sufficiency}, we only leave the selected $r$ most important neurons $\mathcal{N}_{f(x_i)}$ as activated, setting the activation of the remaining neurons $\overline{\mathcal{N}_{f(x_i)}}$ to zero. If the original prediction of the model $f$ is preserved, we claim that the neurons in $\mathcal{N}_{f(x_i)}$ are sufficient to reveal the parametric knowledge employed by $f$ in its original prediction:
\begin{equation}~\label{eq:suff}
\begin{split}
    f(x_i; zero(\overline{\mathcal{N}_{f(x_i)}})) = \widehat{y_i} ; f(x_i) = \widehat{y_i}' \\
    \textrm{If }\widehat{y_i} = \widehat{y_i}'\textrm{, }\mathcal{N}_{f(x_i)}\textrm{ is sufficient.}
\end{split}
\end{equation}
 
We also employ a \textbf{Comprehensiveness} measure to estimate whether the explanation is complete and includes all parametric knowledge that is necessary to understand the model's inner workings. Contrary to Sufficiency, the most important neurons $\mathcal{N}_{f(x_i)}$ are discarded by setting their activation to zero; we then verify whether the prediction changes:
\begin{equation}~\label{eq:compr}
\begin{split}
    f(x_i; zero(\mathcal{N}_{f(x_i)})) = \widehat{y_i} ; f(x_i) = \widehat{y_i}' \\
    \textrm{If }\widehat{y_i} \ne \widehat{y_i}'\textrm{, }\mathcal{N}_{f(x_i)}\textrm{ is comprehensive.}
\end{split}
\end{equation}

For the final Sufficiency and Comprehensiveness measures, we count the number of instances in the test set $X^{Test}$ where Eq. \ref{eq:suff} and Eq. \ref{eq:compr}, correspondingly, hold. 

\subsection{Fine-tuning with Influential Training Instances}
\label{sec:test:ia}
To evaluate the effectiveness of an attribution method in discovering the training instances that affect the learned parametric knowledge by the model, we conduct \textbf{IA Faithfulness} tests. To do so, we fine-tune a model $f$ only with the most influential training instances $\mathcal{S}_i$ and estimate the number of preserved predictions on the test set $X^{Test}$.

\section{Experimental Setup}\label{sec:experiments}

\subsection{Datasets}
In exploring natural language understanding, our research is directed towards tasks that necessitate complex reasoning and engage deeply with a model's acquired task knowledge. Our experiments involve Natural Language Inference (NLI), Fact-Checking (FC) and Question Answering (QA). We focus on selecting both diverse and complex tasks, thus, providing a robust platform for assessing the efficacy of IA and NA methods. One of the selected datasets -- MNLI, is also commonly employed for evaluating attribution methods in related work. For NLI, the Multi-Genre Natural Language Inference (MNLI) is used \cite{williams-etal-2018-mnli}. Due to the computational cost, we sampled 10K training instances from the training dataset (393K in total) following \citet{pezeshkpour-etal-2021-empirical}'s work on the empirical comparison of IA methods. Next, we choose AVeriTeC \cite{schlichtkrull2023averitec}, a highly curated FC dataset. It reflects the characteristics of real-world claims, thus requiring the model to go through more complex reasoning. Finally, to investigate the attribution methods' explanation on a QA task, we work with COmmonSense QA (CoS-QA) \cite{talmor-etal-2019-commonsenseqa}, a multiple-choice Question Answering dataset built upon an external knowledge graph \cite{speer2018conceptnet}.\footnote{Although it is common to provide the external knowledge source to the model for the CoS-QA, we focus on training with the question-answer pairs only as we aim to investigate a model's parametric knowledge.}

\subsection{Models}

We select three open-sourced different autoregressive language models including OPT-125m \cite{zhang2022opt}, BLOOM-560m \cite{yong-etal-2023-bloom} and Pythia-410m \cite{biderman2023pythia}. To see the impact of model size on the attribution explanations, all models are in different sizes. OPT is trained with a de-duplicated version of the Pile dataset \cite{gao2020pile}. For Pythia, we choose the version that is trained with the original Pile dataset, containing duplicated files. BLOOM is oriented to the scientific domain employing multilingual scientific pretraining datasets. The models' performances on the tasks are available in Appendix \ref{sec:appendix:a}. To fine-tune the LMs, we use a Sequence Classification Head on the last token's hidden representation and train the models with cross-entropy loss. We provide further details about fine-tuning in Appendix \ref{sec:appendix:a}.

\subsection{Attribution Methods}

For IA methods, we consider two representative IA methods, IF \cite{pmlr-v70-koh17a} and Gradient Similarity (GS) method \cite{inputsim-nips-2019}. Given a model $f$ and test instance $x_i$, IF (Equation \ref{eq:if}) computes the importance score of training instance $x_{a}^{train}$: $s_{i, a}^{train}$ by upweighting $x_a^{train}$ with the Hessian of the loss function, $\frac{d\epsilon_i}{df} = -H_f^{-1}\nabla_f\mathcal{L}(x_a^{train}, \widehat{y_a}, f)$. We refer to \citet{pmlr-v70-koh17a} for more details. GS (Equation \ref{eq:gs}) takes the dot product of the gradients of $x_i$ and $x_m^{train}$. They are defined as follows:

\begin{equation}
    s_{i, a}^{train}=\nabla\mathcal{L}(x_i, \widehat{y_i}, f)^T \cdot \frac{d\epsilon_i}{df}
    \label{eq:if}
\end{equation}

\begin{equation}
    s_{i, a}^{train}=\nabla_f\mathcal{L}(x_i) \cdot \nabla_f\mathcal{L}(x_a^{train})
    \label{eq:gs}
\end{equation}

Due to the heavy computational cost, the MLP classification layer weight is used to attain the gradients instead of the entire model weights. Prior work showed that the attribution results from the entire weight and the MLP classification layer show a high correlation \cite{pezeshkpour-etal-2021-empirical}.

For NA, we adapt the application of integrated gradients \cite{sundararajan2017axiomatic} for discovering important neurons \cite{dai-etal-2022-knowledge}. The neuron attribution score $ns_i^l$ of $l$\textsuperscript{th} neuron on $f(x_i)$ is: 

\begin{equation}
    ns_i^l = \frac{\overline{w}_l}{m}\sum_{k=1}^m \frac{\partial P_x(\frac{k}{m}\overline{w}_l)}{\partial w_l}
    \label{eq:2}
\end{equation}

where $\overline{w}_l$ is the neuron activation value of the $l$\textsuperscript{th} neuron from the MLP layer. Equation \ref{eq:2} leverages Riemann approximation by scaling the neuron's activation value from 0 to its original value.

\begin{table}[]
\resizebox{\columnwidth}{!}{%
\begin{tabular}{@{}lcrrr@{}}
\toprule
\textbf{Method} & \textbf{Model}              & \multicolumn{1}{c}{\textbf{AVeriTeC}} & \multicolumn{1}{c}{\textbf{MNLI}} & \textbf{CoS-QA} \\ \midrule
Full                                 &                              & 2768                                  & 10000                              & 8767                                \\ \midrule
NA-Instances                                   & \multirow{3}{*}{OPT-125m}    & 427                                   & 795                               & 325                                 \\
IF                                   &                              & 221                                   & 588                               & 2017                                \\
GS                                   &                              & 209                                   & 569                               & 1897                                \\ \midrule
NA-Instances                                 & \multirow{3}{*}{Pythia-410m} & 407                                   & 785                               & 74                                  \\
IF                                   &                              & 108                                   & 102                               & 700                                 \\
GS                                   &                              & 106                                   & 102                               & 700                                 \\ \midrule
NA-Instances                                   & \multirow{3}{*}{BLOOM-560m}  & 246                                   & 1776                              & 324                                 \\
IF                                   &                              & 43                                    & 30                                & 263                                 \\
GS                                   &                              & 43                                    & 30                                & 263\\ \bottomrule
\end{tabular}%
}
\caption{Number of unique instances from the collection of top-10 most influential training instances for each test instance in the corresponding dataset.}
\label{table1}
\end{table}

\subsection{Attribution Method Tests}
For neuron attribution, we employ faithfulness tests covering both the sufficiency and comprehensiveness of the neurons to represent the model's parametric knowledge. For Sufficiency, we take the $r=1$ neuron with the highest attribution score. For Comprehensiveness, $r=100$ neurons are selected. To verify the utility of the neurons selected by the attribution methods, we also choose the same number of neurons randomly and present the result as a \textbf{Random} baseline.

\section{Results}

\begin{table*}[]
\centering
\resizebox{0.6\textwidth}{!}{%
\begin{tabular}{@{}lclrr|rrr@{}}
\toprule
          & \multicolumn{1}{l}{}                                                    & \multicolumn{3}{c|}{\textbf{Sufficiency $\uparrow$}}                                           & \multicolumn{3}{c}{\textbf{Comprehensiveness $\downarrow$}}                                   \\ \midrule
          & \multicolumn{1}{l}{}                                                    & \multicolumn{1}{c}{AVeriTeC} & \multicolumn{1}{l}{MNLI} & \multicolumn{1}{l|}{CoS-QA} & \multicolumn{1}{l}{AVeriTeC} & \multicolumn{1}{l}{MNLI} & \multicolumn{1}{l}{CoS-QA} \\ \midrule
Random    & \multirow{4}{*}{\begin{tabular}[c]{@{}c@{}}OPT-125m\end{tabular}}    & 87.00                        & 88.77                    & 81.93                       & 99.93                        & 99.77                    & 99.34                      \\
NA        &                                                                         & 88.00                        & 88.80                    & 81.90                       & 67.80                        & 94.99                    & 100.0                      \\
IF-Neuron &                                                                         & 87.00                        & 88.11                    & 81.90                       & 98.60                        & 99.76                    & 100.0                      \\
GD-Neuron &                                                                         & 86.40                        & 88.10                    & 81.90                       & 96.20                        & 99.67                    & 100.0                      \\ \midrule
Random    & \multirow{4}{*}{\begin{tabular}[c]{@{}c@{}}Pythia-410m\end{tabular}} & 77.80                        & 86.23                    & 65.30                       & 99.93                        & 99.90                    & 99.65                      \\
NA        &                                                                         & 77.60                        & 86.21                    & 65.27                       & 99.84                        & 99.82                    & 99.84                      \\
IF-Neuron &                                                                         & 77.80                        & 87.73                    & 65.27                       & 100.0                        & 99.91                    & 99.75                      \\
GD-Neuron &                                                                         & 77.60                        & 87.74                    & 65.27                       & 100.0                        & 99.95                    & 99.75
    \\ \midrule
Random    & \multirow{4}{*}{\begin{tabular}[c]{@{}c@{}}BLOOM-560m\end{tabular}}  & 69.80                        & 82.31                    & 75.92                       & 99.80                        & 99.89                    & 99.48                      \\
NA        &                                                                         & 70.60                        & 82.42                    & 76.00                       & 68.80                        & 94.69                    & 99.67                      \\
IF-Neuron &                                                                         & 70.00                        & 81.02                    & 76.17                       & 95.20                        & 99.95                    & 99.75                      \\
GD-Neuron &                                                                         & 70.00                        & 81.01                    & 75.92                       & 88.80                        & 98.85                    & 99.92                     \\ \bottomrule
\end{tabular}%
}\caption{Faithfulness tests (Sufficiency/Comprehensiveness; \S\ref{sec:test:na}), percentage of preserved predictions by keeping/suppressing important neurons for the prediction of each test instance. For Sufficiency, we choose to keep the $r=1$ important neuron activated. For Comprehensiveness, we suppress the $r=100$ most important neurons. The results are averaged over three runs with different seeds. }
\label{table2}
\end{table*}

\subsection{Number of Unique Influential Instances}

Table \ref{table1} showcases the number of unique instances in the collection of the 10 most influential training instances for each test instance. Generally, we observe that \textit{IF and GS usually identify a smaller set of unique instances compared to \nanew} -- e.g., 30 vs 1776 unique instances discovered by IF/GS vs. \nanew\ correspondingly for MNLI and BLOOM-560m. An exception is the CoS-QA dataset, where we hypothesise that the majority of training instances in the CoS-QA dataset have fewer common neurons with the test instances as the test dataset of CoS-QA has fewer concepts in common with the training dataset. Empirically, we find that although most of the concepts\footnote{The CoS-QA dataset provides concept annotations for each instance.} in the test dataset were covered in the training dataset, the concepts from the test dataset only represent a small proportion of all concepts within the training dataset. There were 2099 unique concepts in the training dataset and only 725 concepts were used in the test dataset. This finding obtained from \nanew\ is in line with \citet{huang-etal-2023-rigorously}'s work, which claims that each neuron corresponds to a concept in an instance.  

Further, we observe that \textit{increasing the model size correlates with a decrease in the number of unique training influential instances}. This trend underscores the tendency towards homogeneity in the results produced by IA methods. Moreover, it highlights the potential advantage of the combination of IA and NA methods to retrieve a more diverse set of influential training instances. We further investigate the heterogeneity of the group of influential training instances in \S\ref{sec:analysis}.

\begin{figure*}[t]
\begin{center}
\includegraphics[width=1.0\textwidth]{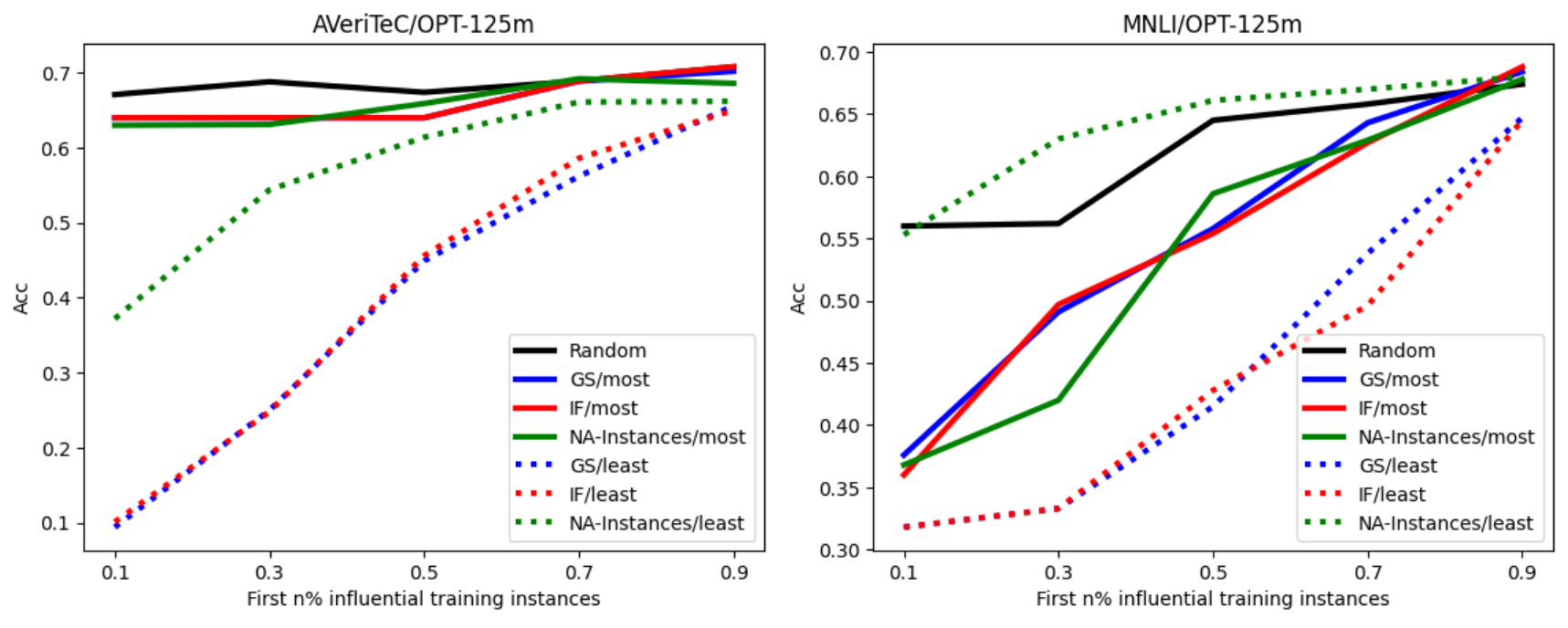}
\caption{Performances with first n\% training instances from each attribution method. For -most methods, top n\% training instances are selected. For -least methods, n\% of negatively influential training instances from the bottom of the list are selected.}
\label{fig2}
\end{center}
\end{figure*}

\subsection{Neuron Attribution Faithfulness Tests}\label{sec:faithfulnessresults}
We evaluate the sufficiency and comprehensiveness of the neurons to represent the model's parametric knowledge. Table \ref{table2} presents the number of the test instances for which the predictions are changed after keeping (Sufficiency) or suppressing (Comprehensiveness) the activation on the selected neurons. We find that NA performs slightly better than other attribution methods for AVeriTeC and MNLI. Furthermore, there are marginal differences between the Random baseline and IF-Neuron/GD-Neuron with less than 1.0 on both Sufficiency and Comprehensiveness. This implies that the \textit{explanations from IF-Neuron/GD-Neuron are not sufficient nor comprehensive enough to reveal the complete parametric knowledge used in the model's prediction}.

Notably, although we suppress the activation of all neurons within the MLP layer except one,\textit{ the model can still recover the original prediction}. We ascribe this phenomenon to the attention layers within the Transformer blocks. This finding aligns with \citet{wiegreffe-pinter-2019-attention}'s argument, that attention weights pose a meaningful impact on the model's prediction and are important for understanding a model's inner workings. For a more holistic understanding of a model's parametric knowledge, we thus suggest that future work also studies attention-based neuron attribution methods.

\subsection{Fine-tuning with Influential Training Instances}\label{sec:finetuningtest}

The models' performances from fine-tuning with different numbers of influential training instances are presented in Figure \ref{fig2}. We also provide the performances on the different datasets and models in Appendix \ref{sec:appendix:b}, Figure \ref{fig6}. From the figure, we find that the accuracies achieved with the first n\% training instances are meaningful enough to show the \textit{different impact on the performance between n\% most influential training instances and n\% least influential training instances}. The biggest gap in the accuracy achieved between training with the most and least influential ones is from the \nanew\ method -- an accuracy gap of 0.6 for the AVeriTeC dataset. However, given that selecting \textit{the same proportion of training instances at random outperforms the attribution methods}, we conclude that the influential training instances selected by IA methods (IF, GS) do not provide any benefit for explaining the performance of the final model. Unexpectedly, the training instances selected by \nanew-least achieve better performance in general than the randomly selected ones on the MNLI dataset. Although \nanew-least shows a different trend on the AVeriTeC and MNLI datasets, it outperforms other least influential groups. Since the group is composed of training instances that have minimal neuron overlapping with the test instances, we attribute this high performance to the instances in the set selected by \nanew-least being more diverse (as seen in general for instances discovered by \nanew\ in Table \ref{table1}) leading to encompassing a more diverse set of the model's parametric knowledge. 

\section{Analysis} \label{sec:analysis}

Next, we investigate what are the characteristics of the group of influential training instances and the group of most important neurons. 

\subsection{Overlap of the Attribution Results}
\label{sec:analysis:1}

\begin{figure}[t]
\includegraphics[width=1.0\columnwidth]{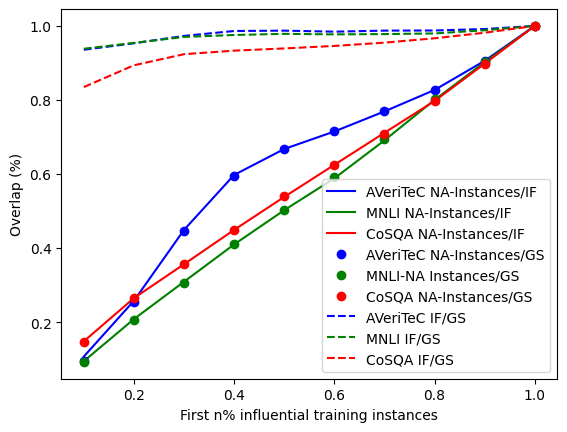}
\caption{\% of training instances at the intersection of the first n\% influential instances discovered by a two of the attribution methods $\in$ \{IF, \nanew, and GS\}.}
\label{fig3}
\end{figure}

\begin{figure}[t]
\includegraphics[width=1.0\columnwidth]{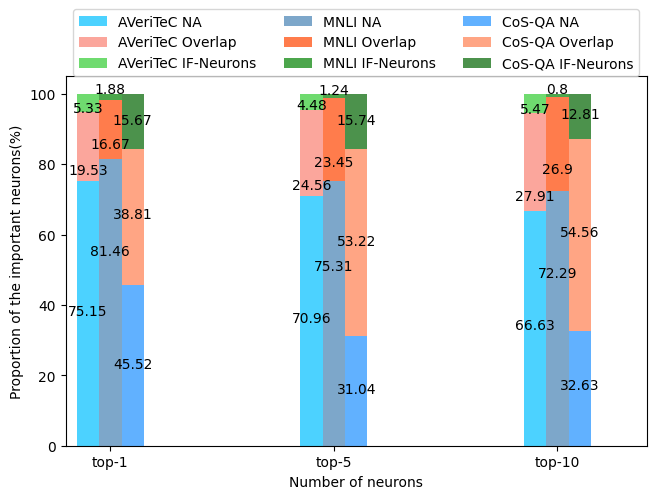}
\caption{\% of the important neurons discovered by NA and IF-Neurons on the union of the top-n important neurons.}
\label{fig4}
\end{figure}

\begin{table*}[h]
\resizebox{\textwidth}{!}{%
\begin{tabular}{@{}lcccc|cccc@{}}
\toprule
                               & \multicolumn{4}{c|}{\textbf{AVeriTeC}}                                                   & \multicolumn{4}{c}{\textbf{MNLI}}                                                        \\ 
\multicolumn{1}{c}{\textbf{}} & \textbf{Cosine Similarity} & \textbf{Loss} & \textbf{Vocabulary} & \textbf{Input Length} & \textbf{Cosine Similarity} & \textbf{Loss} & \textbf{Vocabulary} & \textbf{Input Length} \\ \midrule
Coefficient                    & -1                         & -0.1719       & -0.0018             & 0.036                 & -0.3741                    & -0.3563       & -0.00005            & 0.024                 \\ \midrule
Random                         & 0.300                        & 0.2           & 6977                & 163.1                 & 0.49                       & 0.30          & 6950                & 47.14                 \\
GS-most                        & 0.268                      & 0.3           & 7692                & 197.5                 & 0.61                       & 0.47          & 6427                & 45.98                 \\
IF-most                        & 0.266                      & 0.3           & 7720                & 198.8                 & 0.64                       & 0.38          & 6355                & 45.97                 \\
\nanew-most                        & 0.388                      & 0.2           & 6776.0              & 153.4                 & 0.56                       & 0.52          & 6881                & 45.46                 \\
GS-least                       & 0.278                      & 1.1           & 8199                & 213.5                 & 0.62                       & 0.78          & 6729                & 48.46                 \\
IF-least                       & 0.279                      & 1.1           & 8197                & 211.3                 & 0.62                       & 0.77          & 6838                & 48.42                 \\
\nanew-least                       & 0.245                      & 0.2           & 7978                & 204.1                 & 0.45                       & 0.16          & 6901                & 46.52                 \\ \bottomrule
\end{tabular}%
}
\caption{Diversity analysis on influential training instances discovered for the MNLI and AVeriTeC datasets with the OPT-125m model. Four metrics (Cosine Similarity/Loss/Vocabulary/Input Length; \S\ref{sec:analysis:2}) measure the diversity of the first n\% training instances from each attribution method.}
\label{table3}
\end{table*}

Here we look at the overlap of influential instances as well as the overlap of the important neurons discovered by the corresponding attribution methods. First, we investigate the overlap between the first n influential training instances discovered by IF, \nanew, and GS, which are then used in the evaluation framework for fine-tuning with influential training instances (\S\ref{sec:test:ia}). 
Figure \ref{fig3} shows that for IF and GS, the overlap percentage is high -- greater than 80\%. This also explains their similar performance on the fine-tuning with influential training instances test (\S\ref{sec:finetuningtest}). Furthermore, \textit{compared to the instance attribution methods IF and GS, \nanew\ discovers very different influential instances}. For the first 10\% of most influential instances discovered by each method, we find that \nanew\ and IF or GS have fewer than 20\% instances that are discovered by both methods, which amounts to roughly under 2 influential instances.

Second, we present the proportion of overlapping top-n important neurons selected by NA and IF-Neurons in Figure \ref{fig4}. Results on the overlap of neurons discovered by NA and GS-Neurons show similar trends and can be found in Appendix \ref{sec:appendix:c}, Figure \ref{fig5}. Similar to the diversity of top-n influential training instances, \textit{the proportion of unique important neurons found by NA is again higher than those found by IF-Neurons}. In addition, we find that \textit{most of the neurons found by IF-Neurons are included in the set of NA}. The analytic results from both perspectives underscore the potential of NA methods to reveal the source of the parametric knowledge.

\subsection{Diversity Analysis on the Group of Influential Training Instances}
\label{sec:analysis:2}

From the evaluation results in \S\ref{sec:finetuningtest}, we hypothesize that greater diversity of the influential training instances found by an attribution method yields better performance, which we verify here. 
The heterogeneity of different groups of influential training instances can be measured at the lexical and parametric levels. To estimate lexical diversity, we compute the number of unique tokens (Vocabulary in Table \ref{table3}) from the group of influential training instances and the average length of the training instances (Input Length in Table \ref{table3}) as model input. The cosine similarity between the influential instances with the hidden representations from the last Transformer block (Cosine Similarity in Table \ref{table3}) and the average loss (Loss in Table \ref{table3}) are reported to show the parametric diversity of the selected influential training instances.

Table \ref{table3} presents the result of this analysis on the AVeriTeC dataset and the MNLI dataset with the OPT-125m model, following the previous section. We find that the Random and \nanew-least methods that show a performance of 0.55 accuracy from Figure \ref{fig2} contain more than 6900 unique tokens while other methods with less than 0.40 accuracy have 6600 tokens on average. From the parametric diversity metrics, the methods with lower performance collect training instances with a similar distribution of hidden representations and bigger losses. Furthermore, the least influential training instances discovered by IA methods have higher losses compared to the ones discovered by NA methods. However, we observe that the loss is not an indicator for the most or least influential training instances affecting the model's test set performance from the \nanew\ perspective.

To verify our findings statistically, we implement a simple linear regression model \cite{scikit-learn} that verifies the association between the diversity metrics and the accuracy. Negative coefficients w.r.t. the model's performance show both Cosine Similarity of -0.3741 (MNLI), -1 (AVeriTeC), and Loss of -0.3563 (MNLI), -0.1719 (AVeriTeC). These results support our hypothesis on the \textit{relationship between the diversity of selected training instances and their effectiveness in reaching a high performance with the fine-tuned model}.

\begin{table}[]
\resizebox{\columnwidth}{!}{%
\begin{tabular}{@{}lcc|cc|cc@{}}
\toprule
\multicolumn{1}{c}{\textbf{}} & \multicolumn{2}{c|}{\textbf{OPT-125m}} & \multicolumn{2}{c|}{\textbf{Pythia-410m}} & \multicolumn{2}{c}{\textbf{BLOOM-560m}} \\ 
                               & top-1             & top-10             & top-1              & top-10              & top-1              & top-10              \\ \midrule
Random                         & 0.162             & 0.162              & 0.162              & 0.161               & 0.163              & 0.162               \\
\nanew                             & 0.167             & 0.161              & 0.196              & 0.231               & 0.166              & 0.161               \\
IF                             & 0.159             & 0.137              & 0.182              & 0.182               & 0.111              & 0.131               \\
GS                             & 0.144             & 0.214              & 0.213              & 0.171               & 0.130               & 0.137               \\ \bottomrule
\end{tabular}%
}
\caption{Dataset artifact detection analysis (\S\ref{sec:analysis:3}) presenting the lexical overlap between the premise and the hypothesis from the top-1 and top-10 training instances from the MNLI dataset found to be most influential for the parametric knowledge employed by the model in its prediction in the test instances mispredicted as entailment from the HANS dataset.}
\label{table4}
\end{table}

\subsection{Dataset Artifact Detection}
\label{sec:analysis:3}

For a more human-interpretable analysis of the benefits of each IA method, we conduct a dataset artifact detection analysis with the NLI dataset HANS \cite{mccoy-etal-2019-right}, containing instances representing heuristics that NLI models are likely to learn. 
We take the test instances designed to contain lexical overlap heuristics between the premise and hypothesis to see what types of training instances are found as influential for a model's prediction on the said test instances. We run each instance attribution method on the MNLI training dataset and report the results in Table \ref{table4}.
We find that \textit{our proposed approach \nanew\ generally discovers more training instances which have a higher lexical overlap} rendering the method to perform better in finding artifacts learned by a model from its training dataset.

\section{Conclusion}

In this paper, we propose a unified evaluation framework for comparing and contrasting two different attribution methods, IA and NA. To do so, we first introduce new attribution methods that align the neurons discovered by IA and NA. We assess the sufficiency and comprehensiveness of the explanations for both methods with different faithfulness tests and conduct extensive analyses of their explanations to further investigate the distinct characteristics that yield different results.

Through the proposed evaluation framework, we confirm that IA and NA result in different explanations about the knowledge responsible for the test prediction. This implies the potential advantage of combining IA and NA to unveil a comprehensive view of the LM's parametric knowledge. In addition, the experimental results on the attribution methods' faithfulness suggest that the neurons are not sufficient nor comprehensive enough to fully explain the parametric knowledge used for the test prediction. To complement this drawback of the current attribution methods, we hypothesize that this is due to the importance of the attention weights for encoding knowledge, leaving this exploration for future work. 

\newpage
\section*{Limitations}

Our paper presents an investigation of the knowledge encoded by instance and neuron attribution methods. We perform our experiments using two instance attribution methods, one neuron attribution method, three natural language understanding datasets for three different tasks, and three language models. While we aimed to make a representative selection, future work should investigate other natural language processing tasks not focused on language understanding, as well as more different language models, including very large ones, which we could not study due to computational restrictions. It is also worth noting that the benchmark datasets we used consist of English text only, and we did not study domain adaptation or cross-lingual transfer scenarios. 

One of our core findings was that selecting (in the case of sufficiency) or occluding (for comprehensiveness) knowledge neurons in the MLP layer (see \S\ref{sec:faithfulnessresults}) only tells half the story -- models still perform astonishingly well given only the one most important neuron. This means that, to truly understand the parametric knowledge of LLMs, the interplay between the neurons in the MLP layer and the neurons in the attention heads should be studied.

Lastly, we found that merely using attribution methods to identify instances to train on might not be a good strategy (see \S\ref{sec:finetuningtest}). This is due to the lack of diversity in the resulting training set (see \S\ref{sec:analysis:2}). This naturally begs the question of what the purpose of instance and neuron attribution methods then is. \S\ref{sec:analysis:3} suggests that the answer could be to identify dataset artifacts. However, these phenomena could still be studied in more depth for different downstream tasks.

\section*{Acknowledgements}
$\begin{array}{l}\includegraphics[width=1cm]{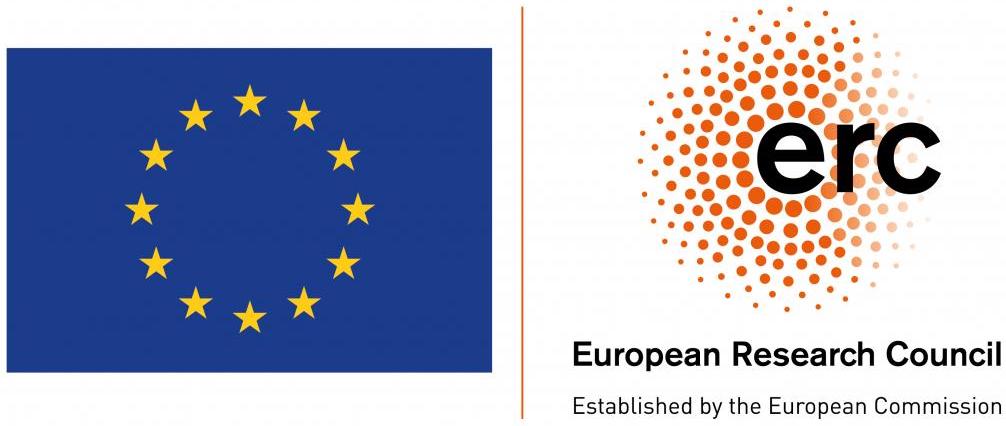} \end{array}$ 
This research was co-funded by the European Union (ERC, ExplainYourself, 101077481), by the Pioneer Centre for AI, DNRF grant number P1, as well as by The Villum Synergy Programme. Views and opinions expressed are however those of the author(s) only and do not necessarily reflect those of the European Union or the European Research Council. Neither the European Union nor the granting authority can be held responsible for them. We thank the anonymous reviewers for their helpful suggestions.

\bibliography{anthology,custom}

\begin{thebibliography}{29}
\expandafter\ifx\csname natexlab\endcsname\relax\def\natexlab#1{#1}\fi

\bibitem[{Bejan et~al.(2023)Bejan, Sokolov, and Filippova}]{bejan-etal-2023-make}
Irina Bejan, Artem Sokolov, and Katja Filippova. 2023.
\newblock \href {https://doi.org/10.18653/v1/2023.emnlp-main.625} {Make every example count: On the stability and utility of self-influence for learning from noisy {NLP} datasets}.
\newblock In \emph{Proceedings of the 2023 Conference on Empirical Methods in Natural Language Processing}, pages 10107--10121, Singapore. Association for Computational Linguistics.

\bibitem[{Biderman et~al.(2023)Biderman, Schoelkopf, Anthony, Bradley, O'Brien, Hallahan, Khan, Purohit, Prashanth, Raff, Skowron, Sutawika, and van~der Wal}]{biderman2023pythia}
Stella Biderman, Hailey Schoelkopf, Quentin Anthony, Herbie Bradley, Kyle O'Brien, Eric Hallahan, Mohammad~Aflah Khan, Shivanshu Purohit, USVSN~Sai Prashanth, Edward Raff, Aviya Skowron, Lintang Sutawika, and Oskar van~der Wal. 2023.
\newblock \href {http://arxiv.org/abs/2304.01373} {Pythia: A suite for analyzing large language models across training and scaling}.

\bibitem[{Charpiat et~al.(2019)Charpiat, Girard, Felardos, and Tarabalka}]{inputsim-nips-2019}
Guillaume Charpiat, Nicolas Girard, Loris Felardos, and Yuliya Tarabalka. 2019.
\newblock \href {https://proceedings.neurips.cc/paper_files/paper/2019/file/c61f571dbd2fb949d3fe5ae1608dd48b-Paper.pdf} {Input similarity from the neural network perspective}.
\newblock In \emph{Advances in Neural Information Processing Systems}, volume~32. Curran Associates, Inc.

\bibitem[{Dai et~al.(2022)Dai, Dong, Hao, Sui, Chang, and Wei}]{dai-etal-2022-knowledge}
Damai Dai, Li~Dong, Yaru Hao, Zhifang Sui, Baobao Chang, and Furu Wei. 2022.
\newblock \href {https://doi.org/10.18653/v1/2022.acl-long.581} {Knowledge neurons in pretrained transformers}.
\newblock In \emph{Proceedings of the 60th Annual Meeting of the Association for Computational Linguistics (Volume 1: Long Papers)}, pages 8493--8502, Dublin, Ireland. Association for Computational Linguistics.

\bibitem[{Fan et~al.(2023)Fan, Dalvi, Durrani, and Sajjad}]{fan2023evaluating}
Yimin Fan, Fahim Dalvi, Nadir Durrani, and Hassan Sajjad. 2023.
\newblock \href {https://openreview.net/forum?id=YiwMpyMdPX} {Evaluating neuron interpretation methods of {NLP} models}.
\newblock In \emph{Thirty-seventh Conference on Neural Information Processing Systems}.

\bibitem[{Gao et~al.(2020)Gao, Biderman, Black, Golding, Hoppe, Foster, Phang, He, Thite, Nabeshima, Presser, and Leahy}]{gao2020pile}
Leo Gao, Stella Biderman, Sid Black, Laurence Golding, Travis Hoppe, Charles Foster, Jason Phang, Horace He, Anish Thite, Noa Nabeshima, Shawn Presser, and Connor Leahy. 2020.
\newblock \href {http://arxiv.org/abs/2101.00027} {The pile: An 800gb dataset of diverse text for language modeling}.

\bibitem[{Garima et~al.(2020)Garima, Liu, Kale, and Sundararajan}]{tracin-nips-2020}
Garima, Frederick Liu, Satyen Kale, and Mukund Sundararajan. 2020.
\newblock Estimating training data influence by tracing gradient descent.
\newblock In \emph{Proceedings of the 34th International Conference on Neural Information Processing Systems}, NIPS'20, Red Hook, NY, USA. Curran Associates Inc.

\bibitem[{Gu et~al.(2023)Gu, Shen, Wang, Wang, Wu, and Mao}]{gu-etal-2023-iaeval}
Peijian Gu, Yaozong Shen, Lijie Wang, Quan Wang, Hua Wu, and Zhendong Mao. 2023.
\newblock \href {https://doi.org/10.18653/v1/2023.findings-emnlp.801} {{IAE}val: A comprehensive evaluation of instance attribution on natural language understanding}.
\newblock In \emph{Findings of the Association for Computational Linguistics: EMNLP 2023}, pages 11966--11977, Singapore. Association for Computational Linguistics.

\bibitem[{Han et~al.(2020)Han, Wallace, and Tsvetkov}]{han-etal-2020-explaining}
Xiaochuang Han, Byron~C. Wallace, and Yulia Tsvetkov. 2020.
\newblock \href {https://doi.org/10.18653/v1/2020.acl-main.492} {Explaining black box predictions and unveiling data artifacts through influence functions}.
\newblock In \emph{Proceedings of the 58th Annual Meeting of the Association for Computational Linguistics}, pages 5553--5563, Online. Association for Computational Linguistics.

\bibitem[{Hase et~al.(2023)Hase, Bansal, Kim, and Ghandeharioun}]{hase2023does}
Peter Hase, Mohit Bansal, Been Kim, and Asma Ghandeharioun. 2023.
\newblock \href {https://openreview.net/forum?id=EldbUlZtbd} {Does localization inform editing? surprising differences in causality-based localization vs. knowledge editing in language models}.
\newblock In \emph{Thirty-seventh Conference on Neural Information Processing Systems}.

\bibitem[{Huang et~al.(2023)Huang, Geiger, D{'}Oosterlinck, Wu, and Potts}]{huang-etal-2023-rigorously}
Jing Huang, Atticus Geiger, Karel D{'}Oosterlinck, Zhengxuan Wu, and Christopher Potts. 2023.
\newblock \href {https://doi.org/10.18653/v1/2023.blackboxnlp-1.24} {Rigorously assessing natural language explanations of neurons}.
\newblock In \emph{Proceedings of the 6th BlackboxNLP Workshop: Analyzing and Interpreting Neural Networks for NLP}, pages 317--331, Singapore. Association for Computational Linguistics.

\bibitem[{J\"{a}rvelin and Kek\"{a}l\"{a}inen(2002)}]{dcg2002acm}
Kalervo J\"{a}rvelin and Jaana Kek\"{a}l\"{a}inen. 2002.
\newblock \href {https://doi.org/10.1145/582415.582418} {Cumulated gain-based evaluation of ir techniques}.
\newblock \emph{ACM Trans. Inf. Syst.}, 20(4):422–446.

\bibitem[{Koh and Liang(2017)}]{pmlr-v70-koh17a}
Pang~Wei Koh and Percy Liang. 2017.
\newblock \href {https://proceedings.mlr.press/v70/koh17a.html} {Understanding black-box predictions via influence functions}.
\newblock In \emph{Proceedings of the 34th International Conference on Machine Learning}, volume~70 of \emph{Proceedings of Machine Learning Research}, pages 1885--1894. PMLR.

\bibitem[{McCoy et~al.(2019)McCoy, Pavlick, and Linzen}]{mccoy-etal-2019-right}
Tom McCoy, Ellie Pavlick, and Tal Linzen. 2019.
\newblock \href {https://doi.org/10.18653/v1/P19-1334} {Right for the wrong reasons: Diagnosing syntactic heuristics in natural language inference}.
\newblock In \emph{Proceedings of the 57th Annual Meeting of the Association for Computational Linguistics}, pages 3428--3448, Florence, Italy. Association for Computational Linguistics.

\bibitem[{Meng et~al.(2022)Meng, Bau, Andonian, and Belinkov}]{meng2022locating}
Kevin Meng, David Bau, Alex~J Andonian, and Yonatan Belinkov. 2022.
\newblock \href {https://openreview.net/forum?id=-h6WAS6eE4} {Locating and editing factual associations in {GPT}}.
\newblock In \emph{Advances in Neural Information Processing Systems}.

\bibitem[{Pedregosa et~al.(2011)Pedregosa, Varoquaux, Gramfort, Michel, Thirion, Grisel, Blondel, Prettenhofer, Weiss, Dubourg, Vanderplas, Passos, Cournapeau, Brucher, Perrot, and Duchesnay}]{scikit-learn}
F.~Pedregosa, G.~Varoquaux, A.~Gramfort, V.~Michel, B.~Thirion, O.~Grisel, M.~Blondel, P.~Prettenhofer, R.~Weiss, V.~Dubourg, J.~Vanderplas, A.~Passos, D.~Cournapeau, M.~Brucher, M.~Perrot, and E.~Duchesnay. 2011.
\newblock Scikit-learn: Machine learning in {P}ython.
\newblock \emph{Journal of Machine Learning Research}, 12:2825--2830.

\bibitem[{Pezeshkpour et~al.(2022)Pezeshkpour, Jain, Singh, and Wallace}]{pezeshkpour-etal-2022-combining}
Pouya Pezeshkpour, Sarthak Jain, Sameer Singh, and Byron Wallace. 2022.
\newblock \href {https://doi.org/10.18653/v1/2022.findings-acl.153} {Combining feature and instance attribution to detect artifacts}.
\newblock In \emph{Findings of the Association for Computational Linguistics: ACL 2022}, pages 1934--1946, Dublin, Ireland. Association for Computational Linguistics.

\bibitem[{Pezeshkpour et~al.(2021)Pezeshkpour, Jain, Wallace, and Singh}]{pezeshkpour-etal-2021-empirical}
Pouya Pezeshkpour, Sarthak Jain, Byron Wallace, and Sameer Singh. 2021.
\newblock \href {https://doi.org/10.18653/v1/2021.naacl-main.75} {An empirical comparison of instance attribution methods for {NLP}}.
\newblock In \emph{Proceedings of the 2021 Conference of the North American Chapter of the Association for Computational Linguistics: Human Language Technologies}, pages 967--975, Online. Association for Computational Linguistics.

\bibitem[{Ross et~al.(2017)Ross, Hughes, and Doshi-Velez}]{Ross2017Right4right}
Andrew~Slavin Ross, Michael~C. Hughes, and Finale Doshi-Velez. 2017.
\newblock Right for the right reasons: training differentiable models by constraining their explanations.
\newblock In \emph{Proceedings of the 26th International Joint Conference on Artificial Intelligence}, IJCAI'17, page 2662–2670. AAAI Press.

\bibitem[{Sajjad et~al.(2022)Sajjad, Durrani, and Dalvi}]{sajjad-etal-2022-neuron}
Hassan Sajjad, Nadir Durrani, and Fahim Dalvi. 2022.
\newblock \href {https://doi.org/10.1162/tacl_a_00519} {Neuron-level interpretation of deep {NLP} models: A survey}.
\newblock \emph{Transactions of the Association for Computational Linguistics}, 10:1285--1303.

\bibitem[{Schlichtkrull et~al.(2023)Schlichtkrull, Guo, and Vlachos}]{schlichtkrull2023averitec}
Michael~Sejr Schlichtkrull, Zhijiang Guo, and Andreas Vlachos. 2023.
\newblock \href {https://openreview.net/forum?id=fKzSz0oyaI} {{AV}eritec: A dataset for real-world claim verification with evidence from the web}.
\newblock In \emph{Thirty-seventh Conference on Neural Information Processing Systems Datasets and Benchmarks Track}.

\bibitem[{Speer et~al.(2018)Speer, Chin, and Havasi}]{speer2018conceptnet}
Robyn Speer, Joshua Chin, and Catherine Havasi. 2018.
\newblock \href {http://arxiv.org/abs/1612.03975} {Conceptnet 5.5: An open multilingual graph of general knowledge}.

\bibitem[{Sundararajan et~al.(2017)Sundararajan, Taly, and Yan}]{sundararajan2017axiomatic}
Mukund Sundararajan, Ankur Taly, and Qiqi Yan. 2017.
\newblock \href {http://arxiv.org/abs/1703.01365} {Axiomatic attribution for deep networks}.

\bibitem[{Talmor et~al.(2019)Talmor, Herzig, Lourie, and Berant}]{talmor-etal-2019-commonsenseqa}
Alon Talmor, Jonathan Herzig, Nicholas Lourie, and Jonathan Berant. 2019.
\newblock \href {https://doi.org/10.18653/v1/N19-1421} {{C}ommonsense{QA}: A question answering challenge targeting commonsense knowledge}.
\newblock In \emph{Proceedings of the 2019 Conference of the North {A}merican Chapter of the Association for Computational Linguistics: Human Language Technologies, Volume 1 (Long and Short Papers)}, pages 4149--4158, Minneapolis, Minnesota. Association for Computational Linguistics.

\bibitem[{Wiegreffe and Pinter(2019)}]{wiegreffe-pinter-2019-attention}
Sarah Wiegreffe and Yuval Pinter. 2019.
\newblock \href {https://doi.org/10.18653/v1/D19-1002} {Attention is not not explanation}.
\newblock In \emph{Proceedings of the 2019 Conference on Empirical Methods in Natural Language Processing and the 9th International Joint Conference on Natural Language Processing (EMNLP-IJCNLP)}, pages 11--20, Hong Kong, China. Association for Computational Linguistics.

\bibitem[{Williams et~al.(2018)Williams, Nangia, and Bowman}]{williams-etal-2018-mnli}
Adina Williams, Nikita Nangia, and Samuel Bowman. 2018.
\newblock \href {https://doi.org/10.18653/v1/N18-1101} {A broad-coverage challenge corpus for sentence understanding through inference}.
\newblock In \emph{Proceedings of the 2018 Conference of the North {A}merican Chapter of the Association for Computational Linguistics: Human Language Technologies, Volume 1 (Long Papers)}, pages 1112--1122, New Orleans, Louisiana. Association for Computational Linguistics.

\bibitem[{Xiong et~al.(2024)Xiong, Li, Zhang, Chen, Sun, Li, Sun, and Du}]{xiong2024explainable}
Haoyi Xiong, Xuhong Li, Xiaofei Zhang, Jiamin Chen, Xinhao Sun, Yuchen Li, Zeyi Sun, and Mengnan Du. 2024.
\newblock \href {http://arxiv.org/abs/2401.04374} {Towards explainable artificial intelligence (xai): A data mining perspective}.

\bibitem[{Yong et~al.(2023)Yong, Schoelkopf, Muennighoff, Aji, Adelani, Almubarak, Bari, Sutawika, Kasai, Baruwa, Winata, Biderman, Raff, Radev, and Nikoulina}]{yong-etal-2023-bloom}
Zheng~Xin Yong, Hailey Schoelkopf, Niklas Muennighoff, Alham~Fikri Aji, David~Ifeoluwa Adelani, Khalid Almubarak, M~Saiful Bari, Lintang Sutawika, Jungo Kasai, Ahmed Baruwa, Genta Winata, Stella Biderman, Edward Raff, Dragomir Radev, and Vassilina Nikoulina. 2023.
\newblock \href {https://doi.org/10.18653/v1/2023.acl-long.653} {{BLOOM}+1: Adding language support to {BLOOM} for zero-shot prompting}.
\newblock In \emph{Proceedings of the 61st Annual Meeting of the Association for Computational Linguistics (Volume 1: Long Papers)}, pages 11682--11703, Toronto, Canada. Association for Computational Linguistics.

\bibitem[{Zhang et~al.(2022)Zhang, Roller, Goyal, Artetxe, Chen, Chen, Dewan, Diab, Li, Lin, Mihaylov, Ott, Shleifer, Shuster, Simig, Koura, Sridhar, Wang, and Zettlemoyer}]{zhang2022opt}
Susan Zhang, Stephen Roller, Naman Goyal, Mikel Artetxe, Moya Chen, Shuohui Chen, Christopher Dewan, Mona Diab, Xian Li, Xi~Victoria Lin, Todor Mihaylov, Myle Ott, Sam Shleifer, Kurt Shuster, Daniel Simig, Punit~Singh Koura, Anjali Sridhar, Tianlu Wang, and Luke Zettlemoyer. 2022.
\newblock \href {http://arxiv.org/abs/2205.01068} {Opt: Open pre-trained transformer language models}.

\end{thebibliography}

\appendix

\section{Implementation Details}
\label{sec:appendix:a}

To adapt the autoregressive LM to the targeted tasks we stack an MLP classification layer with the size of the LM's hidden representation on top of the last token's hidden representation. For AVeriTeC and MNLI, this outputs the probabilities over the labels. For AVeriTeC, we construct an input sequence with a claim and an evidence. With evidence documents that are a pair of question and answer, we concatenate all the pairs. Each LM with MLP classification layer is fine-tuned to predict the label given an input out of \{Supported, Reputed, Conflicting Evidence/Cherrypicking, Not Enough Evidence\}. For the MNLI dataset, a premise and hypothesis are provided as input. The model is trained to predict the label out of \{neutral, contradiction, entailment\}. For the sequences longer than the maximum length of the sequence, we remove tokens from the evidence text.

For the CoS-QA dataset, an input consists of a question and one candidate answer. We forward all possible sequences from one question at the same time to have the model learn the relative difference in the relationship between each candidate's answer and the question. Thus, the model encodes five sequences at the same time and produces five scores. Since we encode each candidate answer individually, the MLP classification layer for CoS-QA outputs a score for each sequence. Then, the model is trained to maximize the score of the sequence that has the correct answer using the listwise loss.

To select the best checkpoint, we train the model for five epochs and report the accuracy on the test dataset from the model checkpoint that shows the best accuracy on the dev dataset. The learning rate is set to $1e-5$ and the maximum length of the sequence is set to 512 tokens for AVeriTeC and MNLI. For the CoS-QA dataset, a maximum of 128 tokens is used. The performances on each dataset can be found in Table \label{5}. For the OPT-125m model, we use a batch size of 8 for AVeriTeC and MNLI, and a batch size of 1 for CoS-QA. For the BLOOM-560m and Pythia-410m models, we use a batch size of 4 for AVeriTeC and MNLI and a batch size of 1 for CoS-QA. The training details are also applied to the fine-tuning of the models for the Instance Attribution Faithfulness Tests.

We use one Titan RTX GPU to fine-tune each model and one A100 GPU to obtain the attribution results.

\begin{table}[]
\resizebox{0.9\columnwidth}{!}{%
\begin{tabular}{@{}ccrr@{}}
\toprule
                                   & \textbf{}                        & \textbf{Dev Acc} & \textbf{Test Acc} \\ \midrule
\multirow{3}{*}{\textbf{AVeriTeC}} & \multicolumn{1}{c|}{OPT-125m}    & 0.75             & 0.72              \\
                                   & \multicolumn{1}{c|}{Pythia-410m} & 0.69             & 0.70               \\
                                   & \multicolumn{1}{c|}{BLOOM-560m}  & 0.72             & 0.72              \\
\multirow{3}{*}{\textbf{MNLI}}     & \multicolumn{1}{c|}{OPT-125m}    & 0.70              & 0.70               \\
                                   & \multicolumn{1}{c|}{Pythia-410m} & 0.63             & 0.63              \\
                                   & \multicolumn{1}{c|}{BLOOM-560m}  & 0.72             & 0.72              \\
\multirow{3}{*}{\textbf{CoS-QA}}   & \multicolumn{1}{c|}{OPT-125m}    & 0.49             & 0.53             \\
                                   & \multicolumn{1}{c|}{Pythia-410m} & 0.42             & 0.44              \\
                                   & \multicolumn{1}{c|}{BLOOM-560m}  & 0.51             & 0.53             \\
\multirow{3}{*}{\textbf{HANS}}     & \multicolumn{1}{c|}{OPT-125m}    & -                & 0.51              \\
                                   & \multicolumn{1}{c|}{Pythia-410m} & -                & 0.50               \\
                                   & \multicolumn{1}{c|}{BLOOM-560m}  & -                & 0.50               \\
                                    \bottomrule
\end{tabular}%
}
\centering
\caption{The performance on each dataset from three different models.}
\label{table5}
\end{table}

\begin{figure}[t]
\includegraphics[width=1.0\columnwidth]{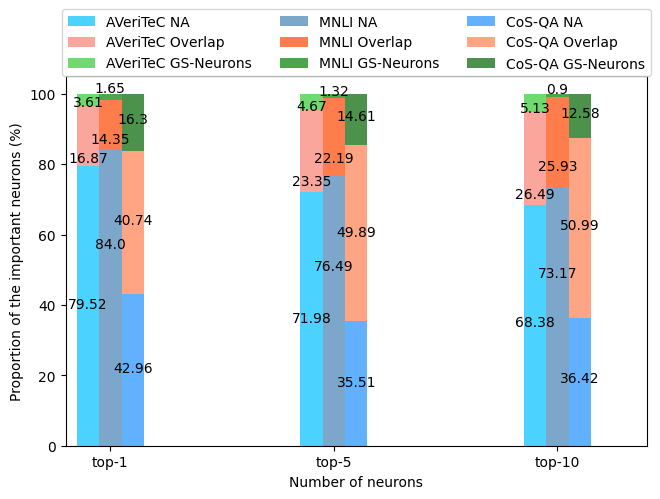}
\caption{\% of the important neurons discovered by NA and GS-Neurons on the union of the top-n important neurons.}
\label{fig5}
\end{figure}

\begin{figure*}[t]
\includegraphics[width=0.9\textwidth]{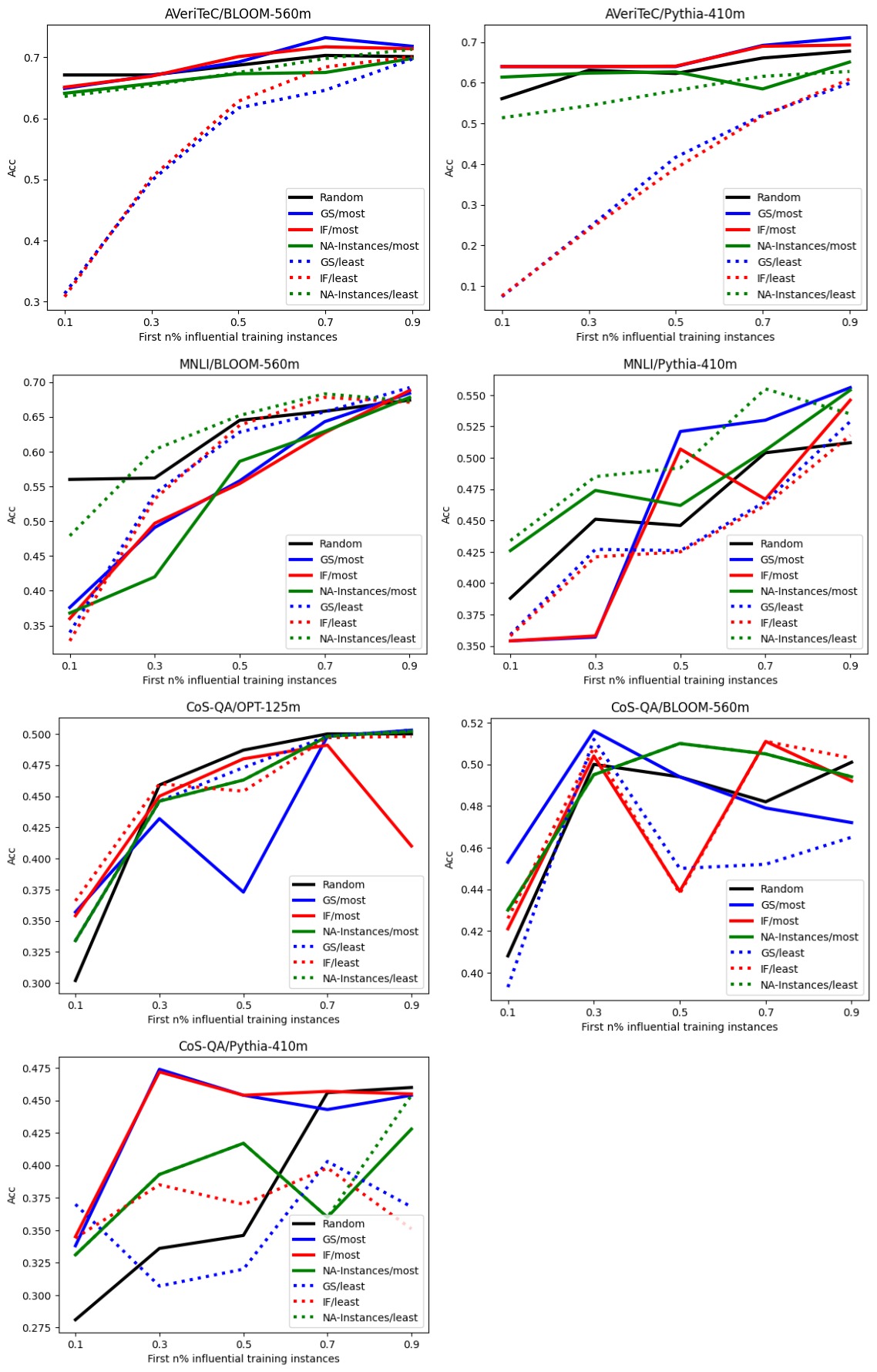}
\centering
\caption{Performances with first n\% training instances from each attribution method. For -most methods, top n\% training instances are selected. For -least methods, n\% of negatively influential training instances from the bottom of the list are selected.}
\label{fig6}
\end{figure*}

\section{Fine-tuning with Influential Training Instances}
\label{sec:appendix:b}

Figure \ref{fig6} presents the models' performances from \S\ref{sec:finetuningtest}. By conducting the evaluation on the sufficiency of the influential training instances, we confirm that various language models show similar trends with the same dataset. For the AVeriTeC, the group of randomly selected training instances (Random) outperforms other groups. On the other hand, the training instances selected by NA-Instances/least consistently show better performances than any other groups in general for the MNLI. With the CoS-QA, we observe that the training instances from -most methods perform better than the group of instances from -least methods.

\section{Overlap among the Group of Important Neurons}
\label{sec:appendix:c}

Figure \ref{fig5} shows the proportion overlapping top-n important neurons selected by NA and GS-Neurons. We refer \S\ref{sec:analysis:1} for details and analysis.

\end{document}